\documentclass[10pt,twocolumn,letterpaper]{article}

\usepackage{cvpr}
\usepackage{times}
\usepackage{epsfig}
\usepackage{graphicx}
\usepackage{amsmath}
\usepackage{amssymb}

\usepackage{xcolor}

\newcommand{\el}[1]{#1~\textit{et al.}}

\usepackage{caption}
\usepackage{subcaption}
\usepackage{siunitx}

\usepackage[breaklinks=true,bookmarks=false]{hyperref}

\cvprfinalcopy 

\def\cvprPaperID{7674} 

\ifcvprfinal\fi
\begin{document}
	
	\title{Towards Rolling Shutter Correction and Deblurring in Dynamic Scenes}
	
	\author{Zhihang Zhong$^{1,2}$\qquad Yinqiang Zheng$^1$\qquad Imari Sato$^{2,1}$\\
		$^1$The University of Tokyo, Japan\qquad
		$^2$National Institute of Informatics, Japan
	}
	
	
	\twocolumn[{%
		\renewcommand\twocolumn[1][]{#1}%
		\maketitle
		\thispagestyle{empty}
		\begin{center}
			\centering
			\includegraphics[width=\textwidth]{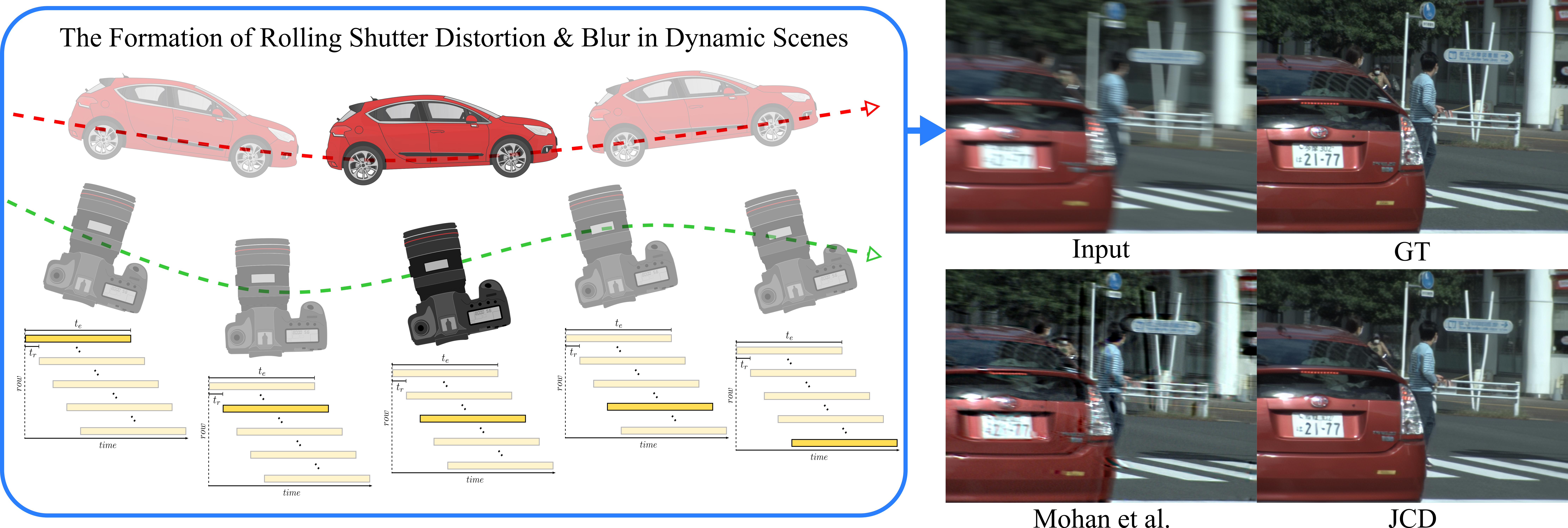}
			\captionof{figure}{\textbf{Joint rolling shutter correction and deblurring (RSCD)}. Image degradation caused by joint rolling shutter distortion and motion blur becomes extremely complex, when both ego-motion and object-motion are involved in dynamic scenes. Existing methods for RSCD such as \el{Mohan}~\cite{mohan2017going} are developed for static scenes. We present the first realistic benchmark dataset (BS-RSCD) and model (JCD) to advance learning-based approaches in this field.}
		\end{center}%
	}]
	
	\begin{abstract}
		Joint rolling shutter correction and deblurring (RSCD) techniques are critical for the prevalent CMOS cameras. However, current approaches are still based on conventional energy optimization and are developed for static scenes. To enable learning-based approaches to address real-world RSCD problem, we contribute the first dataset, BS-RSCD, which includes both ego-motion and object-motion in dynamic scenes. Real distorted and blurry videos with corresponding ground truth are recorded simultaneously via a beam-splitter-based acquisition system.
		
		Since direct application of existing individual rolling shutter correction (RSC) or global shutter deblurring (GSD) methods on RSCD leads to undesirable results due to inherent flaws in the network architecture, we further present the first learning-based model (JCD) for RSCD. The key idea is that we adopt bi-directional warping streams for displacement compensation, while also preserving the non-warped deblurring stream for details restoration. The experimental results demonstrate that JCD achieves state-of-the-art performance on the realistic RSCD dataset (BS-RSCD) and the synthetic RSC dataset (Fastec-RS). The dataset and code are available at \url{https://github.com/zzh-tech/RSCD}\footnote{Correspondence regarding the dataset should be addressed to Y. Zheng (yqzheng@ai.u-tokyo.ac.jp).}.
	\end{abstract}
	
	\section{Introduction}
	Rolling shutter (RS) CMOS cameras dominate the consumer market, especially the mobile phone market, due to their largely reduced power consumption, low cost and compact design~\cite{janesick2009fundamental}. However, if the imaging conditions are not ideal, video recorded through RS mechanism will suffer from compound quality issues. Specifically, the RS distortion (Jello effect) and motion blur become noticeable when there is a large relative motion between the camera and the object, especially in poorly illuminated environment where longer exposure time is required. Joint rolling shutter correction and deblurring (RSCD) techniques are seldom explored and are urgently needed for RS-based devices.
	
	Existing works generally treat rolling shutter correction (RSC) and deblurring as two separate issues. As for RSC methods~\cite{rengarajan2016bows, rengarajan2017unrolling, zhuang2017rolling, lao2018robust, vasu2018occlusion, liu2020deep}, they assume by default that there are no blur effects in the captured image but only distortion caused by the scanning strategy of RS that exposes each row sequentially. The formation of the image $I_r$ with only RS distortion can be described as follows:
	\begin{equation}
		I_r^{(t)}[i] = I_g^{(t - t_m + it_r)}[i],
	\end{equation}
	where $t_m$ equals $(M/2)t_r$; $t_r$ denotes the readout (offset) time for each row of RS; $M$ denotes total number of rows of the image; $I_r^{(t)}[i]$ denotes the $i^{th}$ row of the RS distorted image $I_r$ with the middle moment of exposure at time $t$; $I_g^{(t - t_m + it_r)}[i]$ denotes the same row of the virtual global shutter (GS) image $I_g$ with the middle moment of exposure at time $t - t_m + it_r$, the entire scene of which is captured simultaneously. On the other hand, existing deblurring methods (GSD)~\cite{wulff2014modeling, hyun2015generalized, su2017deep, nah2017deep, tao2018scale, nah2019recurrent, pan2020cascaded, zhong2020efficient} typically assume the target image is captured by GS. Then, the formation of the GS blurry image $I_b$ is given by:
	\begin{equation}
		I_b^{(t)} = \frac{1}{t_e}\int_{t-t_e/2}^{t+t_e/2}I_g^{(t)} dt,
	\end{equation}
	where $t_e$ denotes the exposure time of GS. The combined effect of RS distortion and blur escalates the problem to a new dimension. The formation of RS distorted and blurry image $I_{rb}$ can be described as follows:
	\begin{equation}\label{eq:rscd}
		I_{rb}^{(t)}[i] = \frac{1}{t_e}\int_{t-t_m+it_r-t_e/2}^{t-t_m+it_r+t_e/2}I_g^{(t-t_m+it_r)}[i] dt.
	\end{equation}
	As the reverse process of the Eq.~\eqref{eq:rscd}, RSCD is extremely challenging because it requires estimating pixel-level displacements and blur kernels simultaneously.
	
	Under simplified or idealized conditions, there are few works~\cite{su2015rolling, mohan2017going} that specifically address blind RSCD problem. Assuming a static scene and negligible in-plane rotation, \el{Su}~\cite{su2015rolling} propose RS-BMD to deliver distortion-free and sharp image by estimating parametric trajectory of the camera. \el{Mohan}~\cite{mohan2017going} further remove the limitation of parametric trajectory estimation, allowing their method to handle RS blurry image produced by irregular camera trajectory. However, these methods cannot cope with freely moving objects in dynamic scenes, which are often observed in a real-world scenario. Due to the complexity of relative motion in the RSCD problem, not only is RSCD modeling difficult, but inferring the latent image is also time-consuming.
	
	Recently, the success of deep learning methods and the corresponding large-scale datasets has greatly facilitated the development of image and video restoration techniques. However, the development of learning-based RSCD methods is still hampered by the lack of datasets. Even for pure RSC problem, there is only one public dataset (Fastec-RS~\cite{liu2020deep}) available for data-driven methods, which synthesizes RS images by sequentially copying a row of pixels from captured high-FPS GS images. It still remains a challenge to obtain RS distorted and blurry video and the corresponding GS sharp video for the same scenes in the wild.
	
	In this paper, to enable data-driven methods for RSCD, we propose BS-RSCD, the first dataset used for real-world RSCD task, using a well-designed beam-splitter acquisition system. A RS camera and a GS camera are physically aligned to capture RS distorted and blurry as well as GS sharp video pairs simultaneously. Based on this dataset, we further explore the possibility of applying deep learning to address realistic RSCD problem. The complexity of BS-RSCD brings a new challenge to the existing neural network architecture of GSD or RSC. Through experiments, we found that the existing GSD methods are prone to destroy the original geometric structure of the scene when facing the displacement introduced by RS distortion, while the existing RSC methods are difficult to recover the details from motion blur. To solve this dilemma, we design a novel joint correction and deblurring model (JCD) that incorporates the advantages of both RSC and GSD neural network architectures. To achieve a better trade-off between distortion correction and motion deblurring, JCD fuses bi-directional warped features and non-warped deblurring features at multiple scales using a deformable attention module. Our contributions can be summarized as follows:
	\begin{itemize}
		\item We introduce BS-RSCD, the first dataset for joint RSCD problem in real dynamic scenes, using a beam-splitter acquisition system.
		\item We propose a novel neural network architecture that can handle both RS distortion and blur, using deformable attention module to fuse features from bi-directional warping and deblurring streams.
		\item Experimental results demonstrate the superiority of the proposed method over the state-of-the-art methods in both RSC and RSCD tasks, and show the effectiveness of our BS-RSCD dataset.
	\end{itemize}
	\section{Related works}
	
	
	\subsection{Global Shutter Deblurring}
	
	Early works such as~\el{Fergus}~\cite{fergus2006removing} only consider the uniform blur caused by camera shake. The camera motion is estimated for uniform blur kernel inference. Considering blur of real scenes is spatially varying,~\el{Levin}~\cite{levin2007blind} and~\el{Wulff}~\cite{wulff2014modeling} segment the image into different regions for distinct deconvolution operation with segmentation-wise blur kernel. Then,~\el{Hyun}~\cite{hyun2014segmentation, hyun2015generalized} use energy model to estimate pixel-wise blur kernel without segmentation.~\el{Ren}~\cite{ren2017video} introduce non-linear optical flow for more accurate pixel-wise blur kernel estimation. 
	
	Recently, data-driven approaches achieve leading results for image and video deblurring with an end-to-end manner. Deep learning models are widely used to directly predict sharp image without blur kernel estimation.~\el{Nah}~\cite{nah2017deep} propose a CNN model to remove blur following a coarse-to-fine pyramid architecture.~\el{Tao}~\cite{tao2018scale} further improve the architecture by proposing scale-recurrent structure with ConvLSTM~\cite{shi2015convolutional}. In STRCNN~\cite{su2017deep} and EDVR~\cite{wang2019edvr}, multiple neighboring images of the blurry video are used to provide temporal correlation for deblurring. To better utilize the temporal information, DBN~\cite{hyun2017online}, STFAN~\cite{zhou2019spatio}, IFIRNN~\cite{nah2019recurrent}, ESTRNN~\cite{zhong2020efficient} and CDVD-TSP~\cite{pan2020cascaded} adopt the recurrent structure from RNN and achieve impressive deblurring performance. In addition,~\el{Kupyn}~\cite{Kupyn_2018_CVPR, Kupyn_2019_ICCV} adopt generative adversarial networks to produce smoother and more visually appealing results.
	
	However, the aforementioned image and video deblurring methods have a common implicit assumption that the camera uses a global shutter. These methods will fail when large and complicated displacements caused by the RS distortion appear in the blurry image.
	
	\subsection{Rolling Shutter Correction}
	
	To solve RSC problem,~\el{Forssen}~\cite{forssen2010rectifying} model the camera motion as a parametrised continuous curve and solve parameters using non-linear least squares over inter-frame correspondences.~\el{Grundmann}~\cite{grundmann2012calibration},~\el{Liu}~\cite{liu2013bundled} and~\el{Lao}~\cite{lao2018robust} address the RSC problem with the help of RANSAC~\cite{fischler1981random}.~\el{Rengarajan}~\cite{rengarajan2016bows} correct the RS distorted image with the aid of straight line assumption.~\el{Zhuang}~\cite{zhuang2017rolling} estimate the depth map and motion for RSC by solving a SfM problem using two consecutive RS images.~\el{Vasu}~\cite{vasu2018occlusion} propose a new pipeline to sequentially recover both motion of fast moving camera and scene structure from depth-dependent RS distortion. Recently,~\el{Albl}~\cite{albl2020two} present an simple yet effective method to restore a GS image by using dual RS images with opposite distortion of a same scene.
	
	Inspired by the success of deep learning models, several deep neural network-based methods~\cite{rengarajan2017unrolling, zhuang2019learning, liu2020deep} are proposed and show impressive performance.~\el{Rengarajan}~\cite{rengarajan2017unrolling} propose a CNN model with long rectangular convolutional kernel to estimate the row-wise camera motion for undoing RS distortion.~\el{Zhuang}~\cite{zhuang2019learning} follow the scheme of~\cite{zhuang2017rolling} but use two independent neural network branches to estimate dense depth map and camera motion, respectively.~\el{Liu} first propose an end-to-end network for RSC using a differentiable forward warping module.
	
	Current solutions of the RSC methods cannot handle the blur well presented in RS distorted images even if they are learning-based.
	
	\subsection{Joint Correction and Deblurring}
	Few works consider the challenging situation that joint RS distortion and blur appear in the images simultaneously.~\el{Tourani}~\cite{tourani2016rolling} use feature matches between depth maps to timestamp parametric ego-motion to further achieve RSCD. Their method needs multiple RGBD images as inputs.~\el{Hu}~\cite{hu2016image} use information of inertial sensor in smartphone instead to estimate ego-motion of the RGB images with RS distortion and blur.~\el{Su}~\cite{su2015rolling} first propose an energy model, RS-BMD, to directly estimate the ego-motion without device-specific constraints. However, they discard in-plane rotation to simplify ego-motion.~\el{Mohan}~\cite{mohan2017going} overcome the inability of 3D rotation and irregular ego-motion of RS-BMD. They leverage a generative model for removing RS motion blur and a prior to disambiguate multiple solutions during inversion.
	
	Existing RSCD methods are designed for static scenes with only ego-motion, and there are no learning-based methods for RSCD problem due to the lack of datasets. 
	
	\section{BS-RSCD Dataset}
	
	\begin{figure*}[!h]
		\centering
		\includegraphics[width=\linewidth]{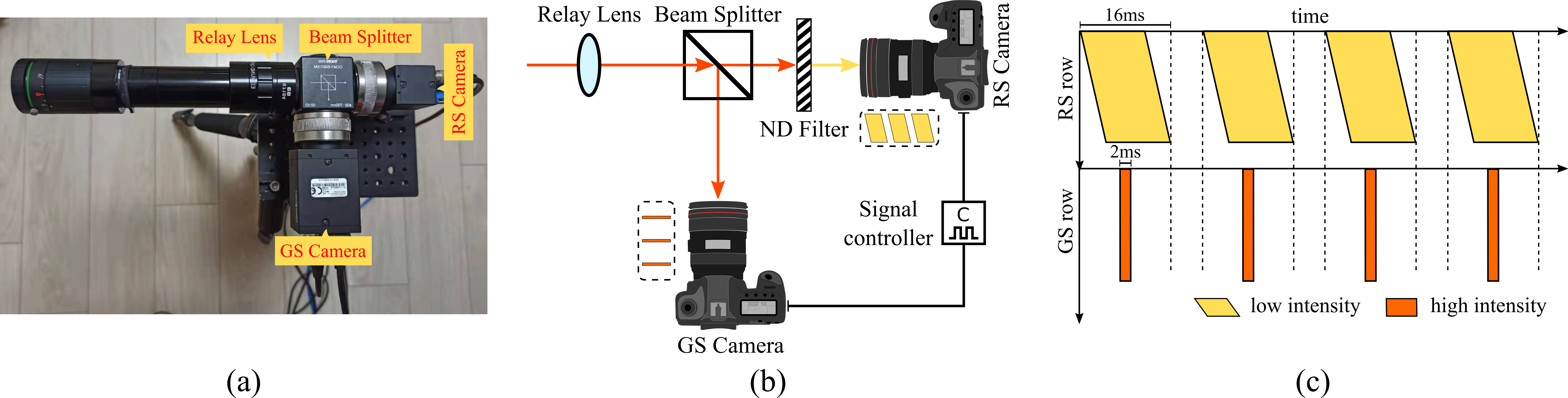}
		\caption{\textbf{Beam-splitter acquisition system.} (a) shows real system used to collect the dataset; (b) is system schematic diagram; (c) is exposure scheme of the system.}
		\label{fig:capture_system}
	\end{figure*}
	
	\begin{figure}[!h]
		\centering
		\includegraphics[width=\linewidth]{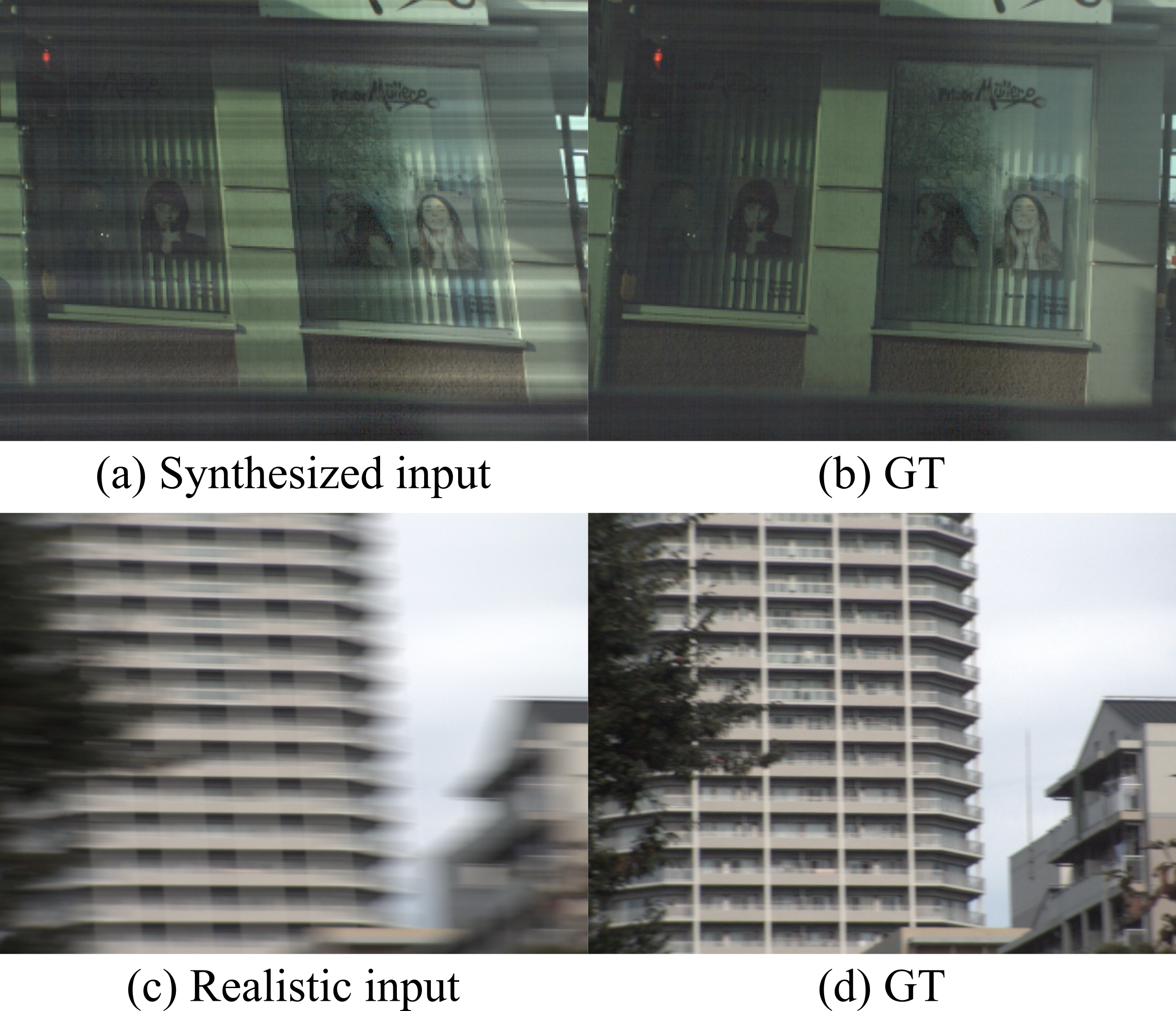}
		\caption{\textbf{ Data samples of Fastec-RS and BS-RSCD.} $1^{st}$ row shows unrealistic case of Fastec-RS with striping artifacts. $2^{nd}$ row shows natural RS distortion and blur in BS-RSCD.}
		\label{fig:dataset_demo}
	\end{figure}
	
	\begin{table}[b]
		\begin{center}
			\begin{tabular}{lccc}
				\hline\hline
				\noalign{\smallskip}
				&  Train & Validation & Test\\
				\hline
				\noalign{\smallskip}
				Sequences& $50$ & $15$ & $15$\\
				Frames/seq.& $50$ & $50$ & $50$\\
				Frames& $2500$  & $750$ & $750$\\
				Resolution& \multicolumn{3}{c}{$640\times480$}\\
				RS camera& \multicolumn{3}{c}{FLIR FL3-U3-13S2C}\\
				GS camera& \multicolumn{3}{c}{FLIR GS3-U3-28S4C}\\
				\hline
			\end{tabular}
		\end{center}
		\caption{\textbf{Configuration of the proposed BS-RSCD.}}
		\label{table:bs_rscd}
	\end{table}
	
	
	It is worth noting that the current mainstream GSD and RSC datasets are synthetic. As for popular GSD datasets, GOPRO~\cite{nah2017deep}, DVD~\cite{su2017deep} and REDS~\cite{nah2019ntire} accumulate consecutive sharp images from high-FPS GS video to generate a blurry image. As for RSC dataset, Fastec-RS~\cite{liu2020deep} uses a GS camera mounted on a ground vehicle to capture high-FPS video with only horizontal motion. Then, RS images are synthesized by sequentially copying a row of pixels from consecutive GS images. Deep learning models trained on the synthetic dataset have limited performance for the data in realistic environment, because the distribution of synthetic data differs from the real data and there may be artifacts. In the case of Fastec-RS, the exposure intensity of the pixel rows from different GS images is inconsistent, resulting in obvious horizontal striping artifacts in the synthetic input, as illustrated in the $1^{st}$ row of Fig.~\ref{fig:dataset_demo}.
	
	To improve the generality of deep learning models, real datasets without synthetic artifacts are essential. Recently, researchers have succeeded in designing specific optical acquisition system to capture real pairs of images or videos for training neural networks.~\el{Zhang}~\cite{zhang2019zoom} use a zoom lens to collect low-resolution and super-resolution image pairs for static scenes.~\el{Rim}~\cite{rim_2020_ECCV} and~\el{Zhong}~\cite{zhong2020efficient} adopt beam-splitter acquisition system to collect single image and video deblurring datasets, respectively. Inspired by the above works, we also propose a beam-splitter acquisition system (Fig.~\ref{fig:capture_system}) to collect the first dataset for RSCD problem, named BS-RSCD. Data samples from the captured video pairs are shown in the $2^{nd}$ row of Fig.~\ref{fig:dataset_demo}. The RS distortion and motion blur in the realistic input look more natural.
	
	\begin{figure*}[h!]
		\centering
		\includegraphics[width=\linewidth]{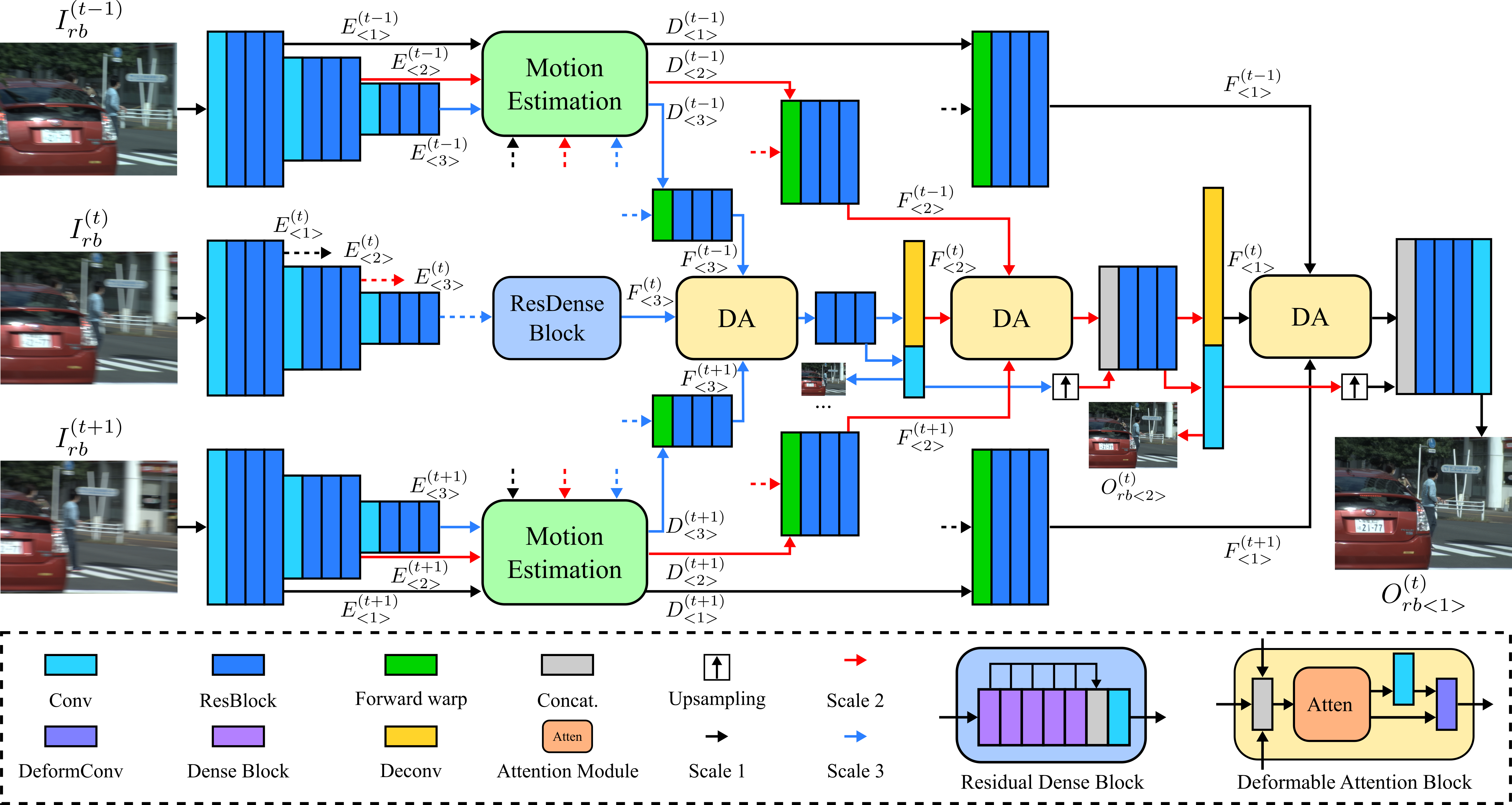}
		\caption{\textbf{Neural network architecture of JCD.} It consists of three streams, including forward and backward warping streams for distortion compensation, and a deblurring stream for detail restoration. A deformable attention block is applied to the concatenation of warped and deblurring features at each scale to predict images in a coarse-to-fine manner.}
		\label{fig:model}
	\end{figure*}
	
	\subsection{Hardware Configuration}
	As for the hardware, we choose a RS camera FLIR FL3-U3-13S2C equipped with a 1/3-inch CMOS sensor, whose pitch size is \SI{3.63}{\micro\metre}. To properly match the sensor specifications of the RS camera, a GS camera FLIR GS3-U3-28S4C is used, whose sensor size is 1/1.8-inch, and pitch size is \SI{3.69}{\micro\metre}. These two cameras are co-axis aligned via a 50/50 beam splitter. As for geometric alignment, the two cameras are first mechanically aligned assisted with collimated laser beams. Later, a homography correction using standard checker pattern is conducted, so as to reduce the alignment error to less than one pixel. The exposure time of the RS and the GS camera is 16ms and 2ms, respectively. Both cameras run at 15fps. We use a wave generator to generate synchronized pulses at 15Hz, and the phase of the pulse for the GS camera is properly delayed, such that the GS exposure time lies in the middle of the RS exposure duration (Fig.~\ref{fig:capture_system}(c)). As for photometric alignment, we put a neutral density filter before the RS camera, such that the intensity values of two cameras are almost equalized. We further use a color checker pattern to correct the RGB response of the GS camera, such that both cameras record the same RGB images, when the scene and cameras are static. 
	
	\subsection{Collected Data}
	
	We collected BS-RSCD in the dynamic urban environment with both ego-motion and object-motion. The configuration of BS-RSCD is listed in Table~\ref{table:bs_rscd}. There are 80 short video sequences of 50 frames each in our dataset (4000 images pairs in total). We divided the BS-RSCD into 50, 15 and 15 sequences representing the training set (2500 image pairs), the validation set (750 image pairs) and the test set (750 image pairs), respectively. The original resolution ($1280\times920$) was downsampled to half ($640\times460$) to suppress the noise level of ground truth.
	
	\section{Method}
	
	Through experiments on BS-RSCD, we found that existing GSD and RSC methods suffer from inherent flaws caused by the design of their neural network architecture, leading to undesirable geometric distortions or blurry details. As for the end-to-end RSC method~\cite{liu2020deep}, latent undistorted image is inferred based on warped features. Learning how to deblur only from warped features is difficult because some of the original information is lost and the process of warping may be imprecise. On the other hand, the encoder-decoder network architecture of the GSD methods fails to preserve the geometry of the objects in the presence of RS distortion. Thus, we propose a novel network for joint correction and deblurring, named JCD, that combines the advantages of RSC and GSD neural network architectures to effectively compensate the displacement and restore the details, shown in Fig.~\ref{fig:model}. The entire structure consists of three streams taking current frame $I_{rb}^{t}$ and two nearby frames $\left\{I_{rb}^{(t-1)}, I_{rb}^{(t+1)}\right\}$ as inputs. The encoder features of the nearby frames are used to produce the warped features of the current frame. In the decoder stage, the features from three streams are fused to predict latent image $O_{rb<1>}^{t}$ from coarse to fine. 
	
	\subsection{Bi-directional Warping Streams and Middle Deblurring Stream}
	First, to promote the precision of warping process, we use bi-directional warping streams to warp different scales of encoder features $\left\{E_{<3>}^{(t)}, E_{<2>}^{(t)}, E_{<1>}^{(t)}\right\}$ of current frame from two directions, forward and backward. We adopt the motion estimation module and differentiable forward warping (DFW) block of~\cite{liu2020deep} to realize both forward warping and backward warping processes. DFW warps features of current frame based on the displacement field ($D_{<s>}^{(t-1)}$ or $D_{<s>}^{(t+1)}$) that is estimated by the correlation layer~\cite{ilg2017flownet} between the same scale features of current frame ($E_{<s>}^{(t)}$) and nearby frame ($E_{<s>}^{(t-1)}$ or $E_{<s>}^{(t+1)}$). The forward warped features $F_{<s>}^{(t-1)}$ of each scale are given by:
	\begin{equation}
		F_{<s>}^{(t-1)} = DFW\left( E_{<s>}^{(t)}, D_{<s>}^{(t-1)} \right), s\in\{1,2,3\}.
	\end{equation}
	Similarly, the backward warped features are given by:
	\begin{equation}
		F_{<s>}^{(t+1)} = DFW\left( E_{<s>}^{(t)}, D_{<s>}^{(t+1)} \right), s\in\{1,2,3\}.
	\end{equation}
	Another advantage of bi-directional warping is that it avoids the need for the model to guess and generate the information of occluded regions around the image borders. Instead, it incorporates the corresponding information into either the forward or backward warped features. In addition to the bi-directional warping streams, we also add an encoder-decoder stream commonly used in deblurring neural networks, denoted as middle deblurring stream. The warping-free features of the middle deblurring stream are more conducive for the model to learn the sharp details of the latent image. The deblurring stream is connected through a residual dense block~\cite{zhang2018residual}, which further extracts rich spatial features $F_{<3>}^{(t)}$ for current frame with different receptive fields.
	
	\subsection{Deformable Attention Module}
	
	In order to effectively integrate the features of bi-directional warping and deblurring streams, we propose a fusion module as deformable attention (DA) block. The DA block handles the concatenation of features ($CAT(F_{<s>}^{(t-1)}, F_{<s>}^{(t)}, F_{<s>}^{(t+1)})$) from different streams via an attention module followed by a deformable convolutional layer~\cite{dai2017deformable, zhu2019deformable}. Channel attention module (squeeze-and-excitation block~\cite{hu2018squeeze}) is adopted to assign channel-wise weight to the concatenated features based on their importance. Since features from warping streams and deblurring stream are not spatially consistent, it is difficult to leverage the information from weighted features using normal 2D convolutional layers. Instead, we use deformable convolution with learnable offset to further fuse the features, which is able to dynamically adjust its receptive field to handle geometric transformation.
	
	\subsection{Loss Function}
	The inference process uses a pyramid structure in a coarse-to-fine manner. There is a predicted result for each scale as $\{O_{rb<3>}^{(t)}, O_{rb<2>}^{(t)}, O_{rb<1>}^{(t)}\}$. The Charbonnier loss and perceptual loss~\cite{johnson2016perceptual} are used to constrain the predicted results at all scales, while total variation loss is applied to the estimated displacement fields to smooth the forward and backward warping processes. The total loss function is given by:
	\begin{equation}
		\mathcal{L}_{total} = \sum_{s=1}^{3}
		\left( \lambda_{c}\mathcal{L}_{c}
		+ \lambda_{p}\mathcal{L}_{p}
		+ \lambda_{tv}\mathcal{L}_{tv} \right).
	\end{equation}
	
	\section{Experiments and Results}
	
	\subsection{Implementation Details}
	
	We implemented our model with PyTorch~\cite{paszke2019pytorch}. The network is trained in $400$ epochs with batch size and learning rate as $8$ and $3\times10^{-4}$. $\lambda_{c}$, $\lambda_{p}$ and $\lambda_{tv}$ are set as $10$, $1$, $0.1$, respectively. Random cropping ($256\times256$) is applied for data augmentation. ADAM~\cite{kingma2014adam} is used to update weight with cosine annealing scheduler. We adopt the deformable convolution layer from Detectron2~\cite{wu2019detectron2} with deformable groups as $8$. Standard metrics PSNR and SSIM, as well as learned perceptual metric LPIPS~\cite{zhang2018perceptual} are used for quantitative evaluation.
	
	\begin{figure*}[t]
		\centering
		\includegraphics[width=\linewidth]{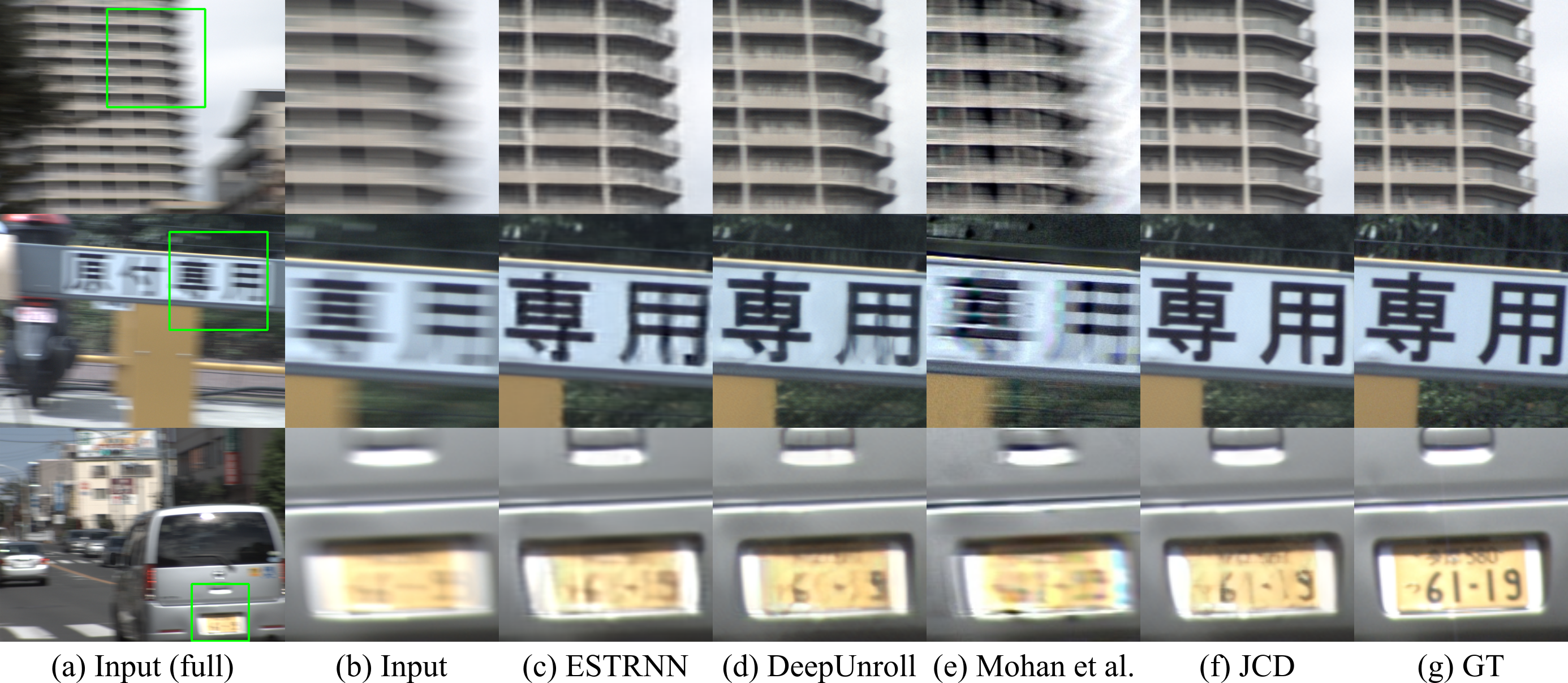}
		\caption{\textbf{Visual results on BS-RSCD dataset.}}
		\label{fig:rscd_comparison}
	\end{figure*}
	
	\begin{figure*}[t]
		\centering
		\includegraphics[width=\linewidth]{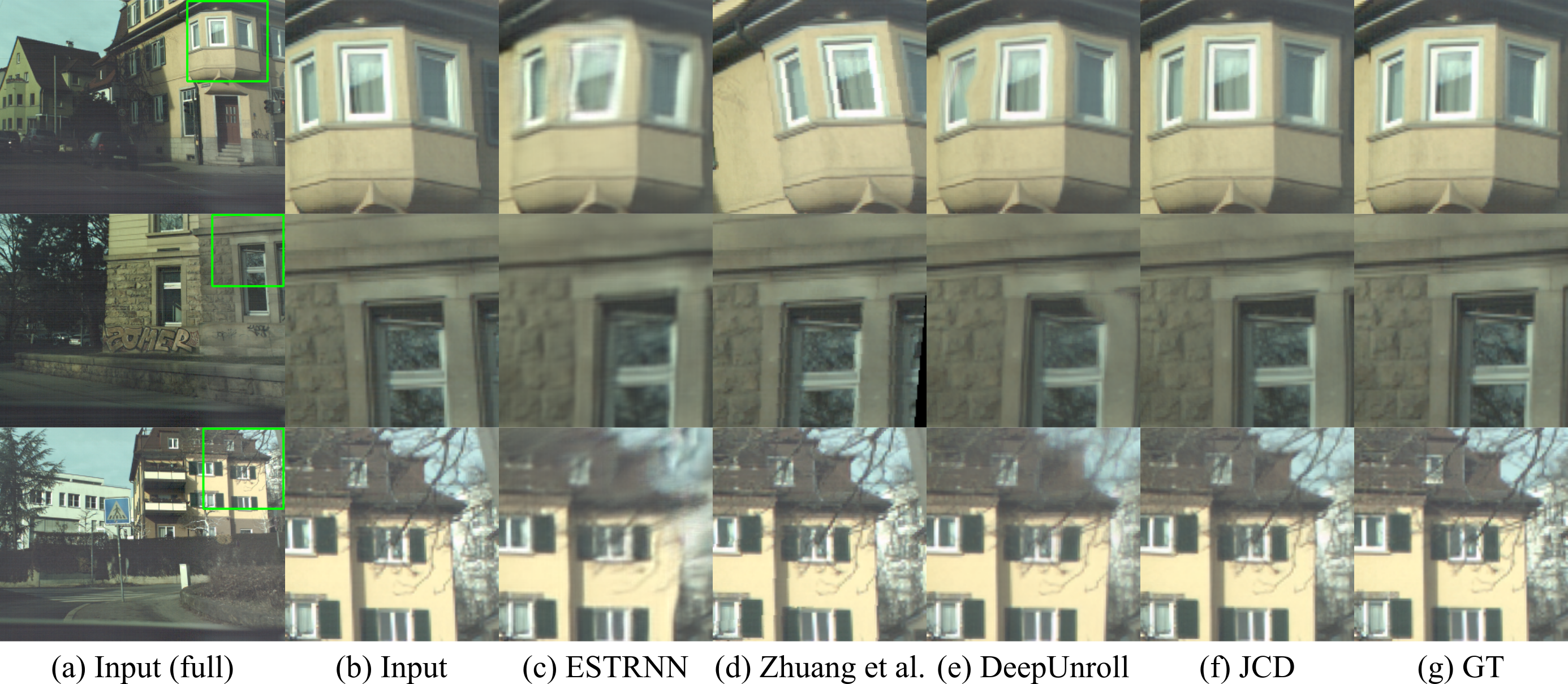}
		\caption{\textbf{Visual results on Fastec-RS dataset.}}
		\label{fig:fastec_rs_comparison}
	\end{figure*}
	
	\begin{table}[!h]
		\begin{center}
			\begin{tabular}{lccc}
				\hline\hline
				\noalign{\smallskip}
				& PSNR $\uparrow$ & SSIM $\uparrow$ & LPIPS $\downarrow$\\
				\hline
				\noalign{\smallskip}
				IFIRNN~\cite{nah2019recurrent}& 25.77 & 0.753 & 0.312\\
				ESTRNN~\cite{zhong2020efficient}& 26.90 & 0.784 & 0.264\\
				DeepUnrollNet~\cite{liu2020deep}& 25.14 & 0.729 & 0.159\\
				RS-BMD~\cite{su2015rolling}& 15.18 & 0.316 & 0.543\\
				\el{Mohan}~\cite{mohan2017going}& 17.46 & 0.420 & 0.404\\
				\hline
				\noalign{\smallskip}
				JCD & 26.42 & 0.757 & 0.122\\
				\hline
			\end{tabular}
		\end{center}
		\caption{\textbf{Quantitative results on BS-RSCD dataset.}}
		\label{table:bs-rscd_comparison}
	\end{table}
	
	
	\begin{table}[!h]
		\begin{center}
			\begin{tabular}{lccc}
				\hline\hline
				\noalign{\smallskip}
				& PSNR $\uparrow$ & SSIM $\uparrow$ & LPIPS $\downarrow$\\
				\hline
				\noalign{\smallskip}
				ESTRNN~\cite{zhong2020efficient}& 27.41 & 0.844 & 0.189\\
				\el{Zhuang}~\cite{zhuang2017rolling}& 21.44 & 0.710 & 0.218\\
				DeepUnrollNet~\cite{liu2020deep}& 27.00 & 0.825 & 0.108\\
				\hline
				\noalign{\smallskip}
				JCD & 24.84 & 0.778 & 0.107 \\
				\hline
			\end{tabular}
		\end{center}
		\caption{\textbf{Quantitative results on Fastec-RS dataset.}}
		\label{table:fastec-rs_comparison}
	\end{table}
	
	
	\begin{table*}[!h]
		\begin{center}
			\begin{tabular}{lcccccc}
				\hline\hline
				\noalign{\smallskip}
				& \multicolumn{3}{c}{BS-RSCD} & \multicolumn{3}{c}{Fastec-RS}\\
				\hline
				\noalign{\smallskip}
				& PSNR $\uparrow$ & SSIM $\uparrow$ & LPIPS $\downarrow$ & PSNR$\uparrow$ & SSIM $\uparrow$ & LPIPS $\downarrow$\\
				\hline
				\noalign{\smallskip}
				JCD (w/o ms) & 25.02 & 0.721 & 0.181 & 25.52 & 0.784 & 0.126 \\
				JCD (w/o dcn) & 25.51 & 0.736 & 0.134 & 25.32 & 0.785 & 0.115 \\
				JCD (w/o bs) & 25.47 & 0.736 & 0.127 & 24.36 & 0.769 & 0.112 \\
				\hline
				\noalign{\smallskip}
				JCD & 26.42 & 0.757 & 0.122 & 24.84 & 0.778 & 0.107 \\
				\hline
			\end{tabular}
		\end{center}
		\caption{\textbf{Quantitative ablation study on BS-RSCD and Fastec-RS.}}
		\label{table:ablation_study}
	\end{table*}
	
	
	\begin{figure*}[!ht]
		\centering
		\includegraphics[width=\linewidth]{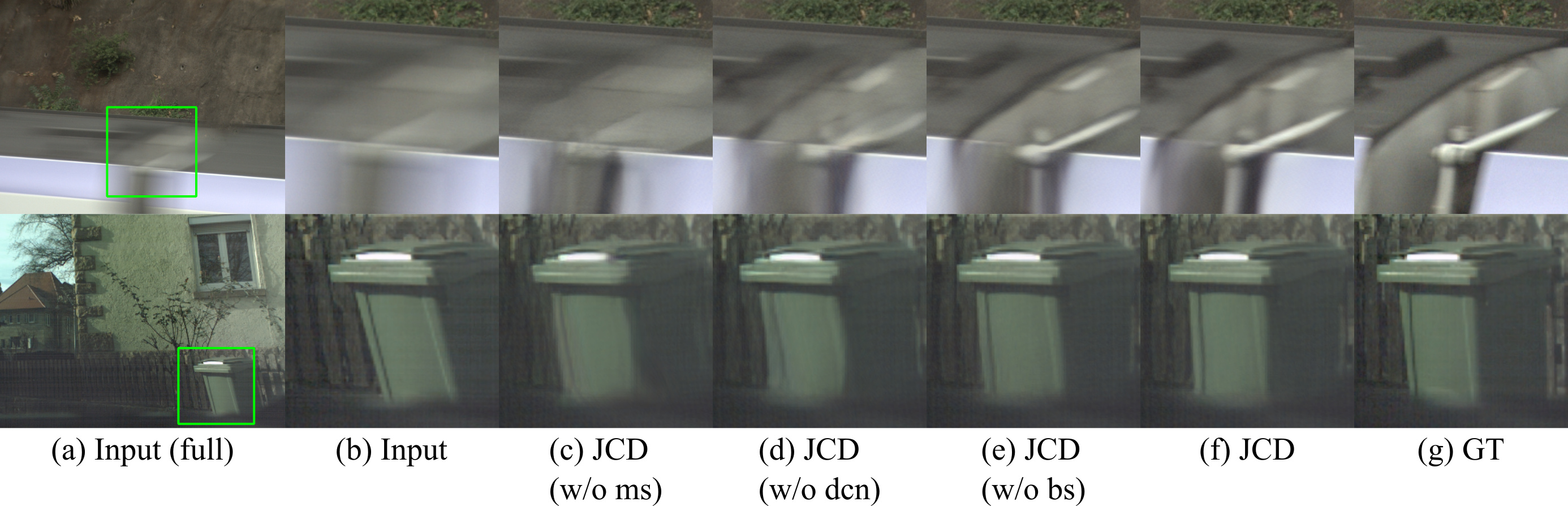}
		\caption{\textbf{Visual results of ablation study.} $1^{st}$ row shows results on BS-RSCD; while $2^{nd}$ row shows results on Fastec-RS.}
		\label{fig:ablation_study}
	\end{figure*}
	
	\begin{figure}[!ht]
		\centering
		\includegraphics[width=\linewidth]{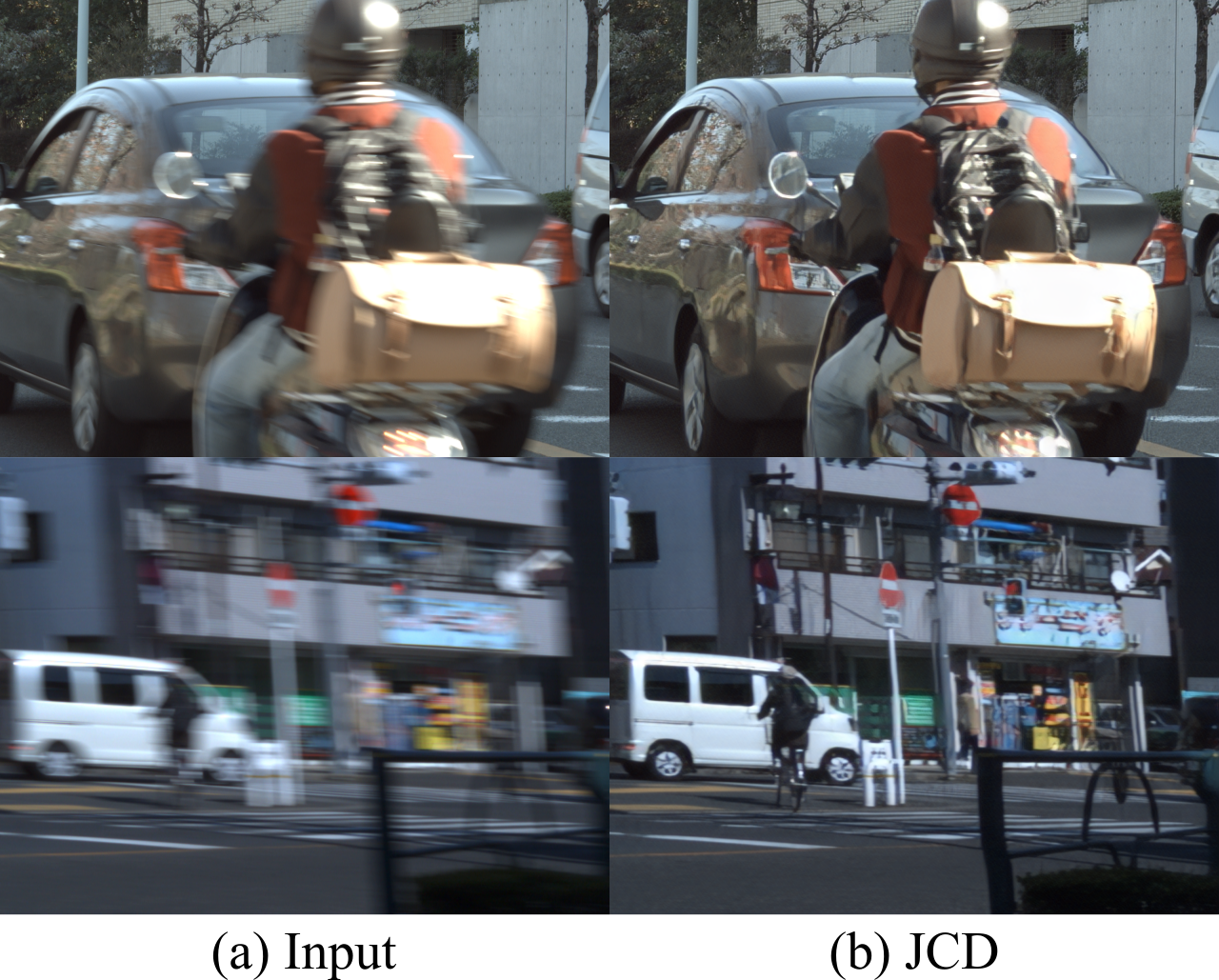}
		\caption{\textbf{Test on the data of third-party cameras.} $1^{st}$ row shows the results of FLIR camera; $2^{nd}$ row shows the results of EO camera.}
		\label{fig:cross_camera_validation}
	\end{figure}
	
	\subsection{Results}
	\subsubsection{Results on BS-RSCD}
	
	We compare JCD against state-of-the-art (SoTA) RSC~\cite{liu2020deep}, GSD~\cite{nah2019recurrent, zhong2020efficient} and RSCD~\cite{su2015rolling, mohan2017going} methods on BS-RSCD. The qualitative results are presented in Fig.~\ref{fig:rscd_comparison}. The predicted results of GSD method, ESTRNN~\cite{zhong2020efficient}, cannot preserve the geometric structure of the object resulting in distorted artifacts. RSC method~\cite{liu2020deep} fails to recover the details of some blurry areas. The performance of SoTA RSCD method is even worse than the learning-based GSD and RSC methods due to the limited modeling capabilities of conventional energy optimization and the assumption of static scenes. Our method achieves the most visually pleasing results, with details and geometry close to the ground truth.
	
	Quantitative comparison is presented in Table~\ref{table:bs-rscd_comparison}. We found that the learned perceptual metrics LPIPS can better reflect the quality of results than SSIM and PSNR for tasks related to RS distortion correction, because LPIPS is less sensitive to pixel position. GSD methods~\cite{nah2019recurrent, zhong2020efficient} without warping process are prone to forcing only some of pixels to the correct position and value, thus cheating non-perceptual metrics (PSNR and SSIM) and obtaining higher values. However, their visual results are much less visually appealing and the LPIPS performance is poor. For example, ESTRNN achieves much higher PSNR ($26.90$ dB) and SSIM ($0.784$) than other methods but only $0.264$ for LPIPS (lower LPIPS is better). Our method achieves best LPIPS performance as $0.122$.
	
	\subsubsection{Results on Fastec-RS}
	
	In addition, we also compare JCD with the SoTA RSC methods~\cite{zhuang2017rolling, liu2020deep} on synthetic RSC dataset, Fastec-RS. The qualitative and quantitative results are presented in Fig.~\ref{fig:fastec_rs_comparison} and Table~\ref{table:fastec-rs_comparison}, respectively. Compared to DeepUnrollNet~\cite{liu2020deep}, our method achieves better visual performance due to more accurate displacement estimation from the bi-directional warping streams and sharper details from the middle deblurring stream. We also test GSD method (ESTRNN~\cite{zhong2020efficient}) on Fastec-RS and observed the same phenomena that the results of ESTRNN have much higher PSNR and SSIM scores but worse visual results and LPIPS score. Compared with DeepUnrollNet, the proposed method has slightly improved in terms of LPIPS, while the improvement in visual performance is more obvious. The visual results of DeepUnrollNet as illustrated in Fig.~\ref{fig:fastec_rs_comparison}(e) exist some undesired distortion and blur.
	
	\subsection{Ablation Study}
	
	We also implemented ablation study experiments to validate the effectiveness of proposed neural network architecture. We implemented other three models including no the middle deblurring stream (w/o ms), no deformable convolution layers (w/o dcn), and only one warping stream (w/o bs). The qualitative and quantitative results on both BS-RSCD and Fastec-RS are presented as Fig.~\ref{fig:ablation_study} and Table~\ref{table:ablation_study}. The results demonstrate that deblurring stream is important for recovering details, the deformable convolutional layers in the fusion module reduce the undesired variations in object geometry, and the bi-directional warping streams further improve the compensation for displacements and details.

	\subsection{Camera Cross Validation}
	
	Finally, we tested the data from two third-party RS cameras, EO 1312LE and FLIR BFLY-PGE-14S2C, using the trained model on BS-RSCD. The visual results are presented in Fig.~\ref{fig:cross_camera_validation}. It demonstrates that model trained on our BS-RSCD is generic cross different devices.
	
	\section{Conclusion}
	In this paper, we constructed the first dataset (BS-RSCD) for joint rolling shutter correction and deblurring task, using a beam-splitter acquisition system. We also proposed the first learning-based method JCD that incorporates the advantages of the network architectures from pure rolling shutter correction and global shutter deblurring tasks. Experimental results demonstrated that fusing the features from both bi-directional warping streams and deblurring stream, using deformable attention module, can restore sharp details and preserve geometric structure of objects in the dynamic scenes. Collecting a larger dataset and proposing a more reasonable and interpretable metrics for the RSCD and RSC tasks will be our future works.
	\section*{Acknowledgement}
	\small{This work was supported in part by the JSPS KAKENHI Grant Numbers JP15H05918 and JP20H05951.}
	
	{\small
		\bibliographystyle{ieee_fullname}
		\bibliography{egbib}

\begin{thebibliography}{10}\itemsep=-1pt

\bibitem{albl2020two}
Cenek Albl, Zuzana Kukelova, Viktor Larsson, Michal Polic, Tomas Pajdla, and
  Konrad Schindler.
\newblock From two rolling shutters to one global shutter.
\newblock In {\em Proceedings of the IEEE/CVF Conference on Computer Vision and
  Pattern Recognition}, pages 2505--2513, 2020.

\bibitem{dai2017deformable}
Jifeng Dai, Haozhi Qi, Yuwen Xiong, Yi Li, Guodong Zhang, Han Hu, and Yichen
  Wei.
\newblock Deformable convolutional networks.
\newblock In {\em Proceedings of the IEEE international conference on computer
  vision}, pages 764--773, 2017.

\bibitem{fergus2006removing}
Rob Fergus, Barun Singh, Aaron Hertzmann, Sam~T Roweis, and William~T Freeman.
\newblock Removing camera shake from a single photograph.
\newblock In {\em ACM SIGGRAPH 2006 Papers}, pages 787--794. 2006.

\bibitem{fischler1981random}
Martin~A Fischler and Robert~C Bolles.
\newblock Random sample consensus: a paradigm for model fitting with
  applications to image analysis and automated cartography.
\newblock {\em Communications of the ACM}, 24(6):381--395, 1981.

\bibitem{forssen2010rectifying}
Per-Erik Forss{\'e}n and Erik Ringaby.
\newblock Rectifying rolling shutter video from hand-held devices.
\newblock In {\em 2010 IEEE Computer Society Conference on Computer Vision and
  Pattern Recognition}, pages 507--514. IEEE, 2010.

\bibitem{grundmann2012calibration}
Matthias Grundmann, Vivek Kwatra, Daniel Castro, and Irfan Essa.
\newblock Calibration-free rolling shutter removal.
\newblock In {\em 2012 IEEE international conference on computational
  photography (ICCP)}, pages 1--8. IEEE, 2012.

\bibitem{hu2018squeeze}
Jie Hu, Li Shen, and Gang Sun.
\newblock Squeeze-and-excitation networks.
\newblock In {\em Proceedings of the IEEE conference on computer vision and
  pattern recognition}, pages 7132--7141, 2018.

\bibitem{hu2016image}
Zhe Hu, Lu Yuan, Stephen Lin, and Ming-Hsuan Yang.
\newblock Image deblurring using smartphone inertial sensors.
\newblock In {\em Proceedings of the IEEE Conference on Computer Vision and
  Pattern Recognition}, pages 1855--1864, 2016.

\bibitem{hyun2014segmentation}
Tae Hyun~Kim and Kyoung Mu~Lee.
\newblock Segmentation-free dynamic scene deblurring.
\newblock In {\em Proceedings of the IEEE Conference on Computer Vision and
  Pattern Recognition}, pages 2766--2773, 2014.

\bibitem{hyun2015generalized}
Tae Hyun~Kim and Kyoung Mu~Lee.
\newblock Generalized video deblurring for dynamic scenes.
\newblock In {\em Proceedings of the IEEE Conference on Computer Vision and
  Pattern Recognition}, pages 5426--5434, 2015.

\bibitem{hyun2017online}
Tae Hyun~Kim, Kyoung Mu~Lee, Bernhard Scholkopf, and Michael Hirsch.
\newblock Online video deblurring via dynamic temporal blending network.
\newblock In {\em Proceedings of the IEEE International Conference on Computer
  Vision}, pages 4038--4047, 2017.

\bibitem{ilg2017flownet}
Eddy Ilg, Nikolaus Mayer, Tonmoy Saikia, Margret Keuper, Alexey Dosovitskiy,
  and Thomas Brox.
\newblock Flownet 2.0: Evolution of optical flow estimation with deep networks.
\newblock In {\em Proceedings of the IEEE conference on computer vision and
  pattern recognition}, pages 2462--2470, 2017.

\bibitem{janesick2009fundamental}
James Janesick, Jeff Pinter, Robert Potter, Tom Elliott, James Andrews, John
  Tower, John Cheng, and Jeanne Bishop.
\newblock Fundamental performance differences between cmos and ccd imagers:
  part iii.
\newblock In {\em Astronomical and Space Optical Systems}, volume 7439, page
  743907. International Society for Optics and Photonics, 2009.

\bibitem{johnson2016perceptual}
Justin Johnson, Alexandre Alahi, and Li Fei-Fei.
\newblock Perceptual losses for real-time style transfer and super-resolution.
\newblock In {\em European conference on computer vision}, pages 694--711.
  Springer, 2016.

\bibitem{kingma2014adam}
Diederik~P Kingma and Jimmy Ba.
\newblock Adam: A method for stochastic optimization.
\newblock {\em arXiv preprint arXiv:1412.6980}, 2014.

\bibitem{Kupyn_2018_CVPR}
Orest Kupyn, Volodymyr Budzan, Mykola Mykhailych, Dmytro Mishkin, and Jiří
  Matas.
\newblock Deblurgan: Blind motion deblurring using conditional adversarial
  networks.
\newblock In {\em Proceedings of the IEEE Conference on Computer Vision and
  Pattern Recognition (CVPR)}, June 2018.

\bibitem{Kupyn_2019_ICCV}
Orest Kupyn, Tetiana Martyniuk, Junru Wu, and Zhangyang Wang.
\newblock Deblurgan-v2: Deblurring (orders-of-magnitude) faster and better.
\newblock In {\em Proceedings of the IEEE/CVF International Conference on
  Computer Vision (ICCV)}, October 2019.

\bibitem{lao2018robust}
Yizhen Lao and Omar Ait-Aider.
\newblock A robust method for strong rolling shutter effects correction using
  lines with automatic feature selection.
\newblock In {\em Proceedings of the IEEE Conference on Computer Vision and
  Pattern Recognition}, pages 4795--4803, 2018.

\bibitem{levin2007blind}
Anat Levin.
\newblock Blind motion deblurring using image statistics.
\newblock In {\em Advances in Neural Information Processing Systems}, pages
  841--848, 2007.

\bibitem{liu2020deep}
Peidong Liu, Zhaopeng Cui, Viktor Larsson, and Marc Pollefeys.
\newblock Deep shutter unrolling network.
\newblock In {\em Proceedings of the IEEE/CVF Conference on Computer Vision and
  Pattern Recognition}, pages 5941--5949, 2020.

\bibitem{liu2013bundled}
Shuaicheng Liu, Lu Yuan, Ping Tan, and Jian Sun.
\newblock Bundled camera paths for video stabilization.
\newblock {\em ACM Transactions on Graphics (TOG)}, 32(4):1--10, 2013.

\bibitem{mohan2017going}
Mahesh~MR Mohan, AN Rajagopalan, and Gunasekaran Seetharaman.
\newblock Going unconstrained with rolling shutter deblurring.
\newblock In {\em Proceedings of the IEEE International Conference on Computer
  Vision}, pages 4010--4018, 2017.

\bibitem{nah2019ntire}
Seungjun Nah, Sungyong Baik, Seokil Hong, Gyeongsik Moon, Sanghyun Son, Radu
  Timofte, and Kyoung Mu~Lee.
\newblock Ntire 2019 challenge on video deblurring and super-resolution:
  Dataset and study.
\newblock In {\em Proceedings of the IEEE Conference on Computer Vision and
  Pattern Recognition Workshops}, pages 0--0, 2019.

\bibitem{nah2017deep}
Seungjun Nah, Tae Hyun~Kim, and Kyoung Mu~Lee.
\newblock Deep multi-scale convolutional neural network for dynamic scene
  deblurring.
\newblock In {\em Proceedings of the IEEE conference on computer vision and
  pattern recognition}, pages 3883--3891, 2017.

\bibitem{nah2019recurrent}
Seungjun Nah, Sanghyun Son, and Kyoung~Mu Lee.
\newblock Recurrent neural networks with intra-frame iterations for video
  deblurring.
\newblock In {\em Proceedings of the IEEE/CVF Conference on Computer Vision and
  Pattern Recognition}, pages 8102--8111, 2019.

\bibitem{pan2020cascaded}
Jinshan Pan, Haoran Bai, and Jinhui Tang.
\newblock Cascaded deep video deblurring using temporal sharpness prior.
\newblock In {\em Proceedings of the IEEE/CVF Conference on Computer Vision and
  Pattern Recognition}, pages 3043--3051, 2020.

\bibitem{paszke2019pytorch}
Adam Paszke, Sam Gross, Francisco Massa, Adam Lerer, James Bradbury, Gregory
  Chanan, Trevor Killeen, Zeming Lin, Natalia Gimelshein, Luca Antiga, et~al.
\newblock Pytorch: An imperative style, high-performance deep learning library.
\newblock In {\em Advances in neural information processing systems}, pages
  8026--8037, 2019.

\bibitem{ren2017video}
Wenqi Ren, Jinshan Pan, Xiaochun Cao, and Ming-Hsuan Yang.
\newblock Video deblurring via semantic segmentation and pixel-wise non-linear
  kernel.
\newblock In {\em Proceedings of the IEEE International Conference on Computer
  Vision}, pages 1077--1085, 2017.

\bibitem{rengarajan2017unrolling}
Vijay Rengarajan, Yogesh Balaji, and AN Rajagopalan.
\newblock Unrolling the shutter: Cnn to correct motion distortions.
\newblock In {\em Proceedings of the IEEE Conference on computer Vision and
  Pattern Recognition}, pages 2291--2299, 2017.

\bibitem{rengarajan2016bows}
Vijay Rengarajan, Ambasamudram~N Rajagopalan, and Rangarajan Aravind.
\newblock From bows to arrows: Rolling shutter rectification of urban scenes.
\newblock In {\em Proceedings of the IEEE Conference on Computer Vision and
  Pattern Recognition}, pages 2773--2781, 2016.

\bibitem{rim_2020_ECCV}
Jaesung Rim, Haeyun Lee, Jucheol Won, and Sunghyun Cho.
\newblock Real-world blur dataset for learning and benchmarking deblurring
  algorithms.
\newblock In {\em Proceedings of the European Conference on Computer Vision
  (ECCV)}, 2020.

\bibitem{shi2015convolutional}
Xingjian Shi, Zhourong Chen, Hao Wang, Dit-Yan Yeung, Wai-Kin Wong, and
  Wang-chun Woo.
\newblock Convolutional lstm network: A machine learning approach for
  precipitation nowcasting.
\newblock {\em Advances in neural information processing systems}, 28:802--810,
  2015.

\bibitem{su2017deep}
Shuochen Su, Mauricio Delbracio, Jue Wang, Guillermo Sapiro, Wolfgang Heidrich,
  and Oliver Wang.
\newblock Deep video deblurring for hand-held cameras.
\newblock In {\em Proceedings of the IEEE Conference on Computer Vision and
  Pattern Recognition}, pages 1279--1288, 2017.

\bibitem{su2015rolling}
Shuochen Su and Wolfgang Heidrich.
\newblock Rolling shutter motion deblurring.
\newblock In {\em Proceedings of the IEEE Conference on Computer Vision and
  Pattern Recognition}, pages 1529--1537, 2015.

\bibitem{tao2018scale}
Xin Tao, Hongyun Gao, Xiaoyong Shen, Jue Wang, and Jiaya Jia.
\newblock Scale-recurrent network for deep image deblurring.
\newblock In {\em Proceedings of the IEEE Conference on Computer Vision and
  Pattern Recognition}, pages 8174--8182, 2018.

\bibitem{tourani2016rolling}
Siddharth Tourani, Sudhanshu Mittal, Akhil Nagariya, Visesh Chari, and Madhava
  Krishna.
\newblock Rolling shutter and motion blur removal for depth cameras.
\newblock In {\em 2016 IEEE international conference on robotics and automation
  (ICRA)}, pages 5098--5105. IEEE, 2016.

\bibitem{vasu2018occlusion}
Subeesh Vasu, Mahesh~MR Mohan, and AN Rajagopalan.
\newblock Occlusion-aware rolling shutter rectification of 3d scenes.
\newblock In {\em Proceedings of the IEEE Conference on Computer Vision and
  Pattern Recognition}, pages 636--645, 2018.

\bibitem{wang2019edvr}
Xintao Wang, Kelvin~CK Chan, Ke Yu, Chao Dong, and Chen Change~Loy.
\newblock Edvr: Video restoration with enhanced deformable convolutional
  networks.
\newblock In {\em Proceedings of the IEEE Conference on Computer Vision and
  Pattern Recognition Workshops}, 2019.

\bibitem{wu2019detectron2}
Yuxin Wu, Alexander Kirillov, Francisco Massa, Wan-Yen Lo, and Ross Girshick.
\newblock Detectron2.
\newblock \url{https://github.com/facebookresearch/detectron2}, 2019.

\bibitem{wulff2014modeling}
Jonas Wulff and Michael~Julian Black.
\newblock Modeling blurred video with layers.
\newblock In {\em European Conference on Computer Vision}, pages 236--252.
  Springer, 2014.

\bibitem{zhang2018perceptual}
Richard Zhang, Phillip Isola, Alexei~A Efros, Eli Shechtman, and Oliver Wang.
\newblock The unreasonable effectiveness of deep features as a perceptual
  metric.
\newblock In {\em CVPR}, 2018.

\bibitem{zhang2019zoom}
Xuaner Zhang, Qifeng Chen, Ren Ng, and Vladlen Koltun.
\newblock Zoom to learn, learn to zoom.
\newblock In {\em Proceedings of the IEEE Conference on Computer Vision and
  Pattern Recognition}, pages 3762--3770, 2019.

\bibitem{zhang2018residual}
Yulun Zhang, Yapeng Tian, Yu Kong, Bineng Zhong, and Yun Fu.
\newblock Residual dense network for image super-resolution.
\newblock In {\em Proceedings of the IEEE conference on computer vision and
  pattern recognition}, pages 2472--2481, 2018.

\bibitem{zhong2020efficient}
Zhihang Zhong, Ye Gao, Yinqiang Zheng, and Bo Zheng.
\newblock Efficient spatio-temporal recurrent neural network for video
  deblurring.
\newblock In {\em European Conference on Computer Vision}, pages 191--207.
  Springer, 2020.

\bibitem{zhou2019spatio}
Shangchen Zhou, Jiawei Zhang, Jinshan Pan, Haozhe Xie, Wangmeng Zuo, and Jimmy
  Ren.
\newblock Spatio-temporal filter adaptive network for video deblurring.
\newblock In {\em Proceedings of the IEEE International Conference on Computer
  Vision}, pages 2482--2491, 2019.

\bibitem{zhu2019deformable}
Xizhou Zhu, Han Hu, Stephen Lin, and Jifeng Dai.
\newblock Deformable convnets v2: More deformable, better results.
\newblock In {\em Proceedings of the IEEE Conference on Computer Vision and
  Pattern Recognition}, pages 9308--9316, 2019.

\bibitem{zhuang2017rolling}
Bingbing Zhuang, Loong-Fah Cheong, and Gim Hee~Lee.
\newblock Rolling-shutter-aware differential sfm and image rectification.
\newblock In {\em Proceedings of the IEEE International Conference on Computer
  Vision}, pages 948--956, 2017.

\bibitem{zhuang2019learning}
Bingbing Zhuang, Quoc-Huy Tran, Pan Ji, Loong-Fah Cheong, and Manmohan
  Chandraker.
\newblock Learning structure-and-motion-aware rolling shutter correction.
\newblock In {\em Proceedings of the IEEE Conference on Computer Vision and
  Pattern Recognition}, pages 4551--4560, 2019.

\end{thebibliography}
	}
	
\end{document}